\PassOptionsToPackage{unicode}{hyperref}
\PassOptionsToPackage{hyphens}{url}

\documentclass{edm_article}

\providecommand{\tightlist}{%
  \setlength{\itemsep}{0pt}\setlength{\parskip}{0pt}}

\usepackage{enumitem}
\setlist{nosep}

\IfFileExists{bookmark.sty}{\usepackage{bookmark}}{\usepackage{hyperref}}
\IfFileExists{xurl.sty}{\usepackage{xurl}}{} 
\hypersetup{
  pdftitle={Towards a Unified Framework for Evaluating Explanations},
    pdfauthor={true, true},
  pdfkeywords={Explainable AI, evaluating explanations, model
transparency, interpretable neural networks},
  hidelinks,
  pdfcreator={LaTeX via pandoc}}

\begin{document}

\title{Towards a Unified Framework for Evaluating Explanations}

\numberofauthors{3}

\author{
\alignauthor
Juan D. Pinto\\
    \affaddr{University of Illinois Urbana-Champaign}\\
    \email{jdpinto2@illinois.edu}
\alignauthor
Luc Paquette\\
    \affaddr{University of Illinois Urbana-Champaign}\\
    \email{lpaq@illinois.edu}
}

\maketitle

\begin{abstract}
The challenge of creating interpretable models has been taken up by two
main research communities: ML researchers primarily focused on
lower-level explainability methods that suit the needs of engineers, and
HCI researchers who have more heavily emphasized user-centered
approaches often based on participatory design methods. This paper
reviews how these communities have evaluated interpretability,
identifying overlaps and semantic misalignments. We propose moving
towards a unified framework of evaluation criteria and lay the
groundwork for such a framework by articulating the relationships
between existing criteria. We argue that explanations serve as mediators
between models and stakeholders, whether for intrinsically interpretable
models or opaque black-box models analyzed via post-hoc techniques. We
further argue that useful explanations require both faithfulness and
intelligibility. Explanation plausibility is a prerequisite for
intelligibility, while stability is a prerequisite for explanation
faithfulness. We illustrate these criteria, as well as specific
evaluation methods, using examples from an ongoing study of an
interpretable neural network for predicting a particular learner
behavior.
\end{abstract}


\section{Introduction}\label{introduction}

The growing awareness in educational data mining (EDM) of a need for
more explainable AI (XAI) has led to the increasing discussion and
adoption of interpretability methods
\cite{khosraviExplainableArtificialIntelligence2022}. Such methods are continually being developed and refined
within research communities such as machine learning (ML) and
human-computer interaction (HCI). However, the critical task of
evaluating the efficacy of the explanations created has not been
sufficiently explored. Tellingly, a systematic review of explainable
student performance prediction models did not find a single study that
evaluated the explanations it produced
\cite{alamriExplainableStudentPerformance2021}. Furthermore, a standardized framework for conducting
such evaluations is still lacking
\cite{liuApplicationsExplainableAIinpress}.

In this position paper, we aim to foster discussion to begin addressing
this gap by proposing the goal of a unified framework for evaluating
explanations. We review concepts critical to the goals of XAI, including
the intended contexts of an explanation, the multiple research milieux
of explainability and intelligibility, and proposed evaluation methods.
We argue that the evaluation of explanations should be based on a set of
criteria that must be met for an explanation to be useful and propose a
hierarchy of criteria that brings together some previously described in
the literature. Finally, we illustrate these criteria using an ongoing
study of an interpretable neural network for predicting a particular
learner behavior. We conclude by discussing the implications of this
initial perspective on an evaluation framework for future research in
XAI.

\section{What to evaluate?}\label{what-to-evaluate}

The explainability literature has often highlighted the difference
between intrinsically interpretable models that are designed with
transparency in mind and opaque black-box models that require post-hoc
explainability methods
\cite{kumarOverviewExplainableAI2023}.
At face value, models and explanations seem like two very different
objects to evaluate. However, even intrinsically interpretable
models---such as linear regression models or decision trees---require
some form of explanation to serve as mediator between the model's
internal state and a user's understanding of it. This is true in cases
of local explainability---eg. the importance of a specific feature in a
decision tree for a particular prediction---but also when global
explainability is the goal---eg. the coefficients of a linear model,
along with their meanings and interactions, which provide an overall
picture of the model's behavior. From this perspective, the evaluation
of explainability can always be treated as the evaluation of
explanations.

\section{Intended context}\label{intended-context}

When evaluating any explanation, one critical aspect to be considered is
the context in which the explanation is to be used. An explanation that
is useful for researchers carefully analyzing and modifying a model's
behavior in a controlled environment may not be useful for a teacher
trying to understand in real time why a student is struggling with a
particular concept. The intended users for which the explanation has
been designed must clearly dictate the criteria used to evaluate it.
This is often what is meant by the term ``human-centered'', which is
used in a commendable effort to distance research from simplistic
technocentric approaches, instead emphasizing the importance of people.
But identifying ``humans'' as the target of our XAI efforts is still far
too broad.

When considering the requirements that an explanation should aim to
fulfill, it is useful to examine both the \emph{knowledge} and
\emph{objectives} of the intended users
\cite{sureshExpertiseRolesFramework2021}.
Teachers, for example, may wish to help specific students with the
insights gained from an explanation. Their knowledge includes their
familiarity with their students and their knowledge of the subject
matter. Students, on the other hand, may wish to know why a learning
platform is making a specific suggestion in order to gauge its
effectiveness. Their knowledge might include their level of familiarity
with self-regulation strategies, their current understanding of the
subject, and clues from what their peers are doing. Researchers may wish
instead to use an explanation to better understand how to improve the
model, which can involve reducing bias, improving performance, or
identifying and fixing bugs that may be present
\cite{vaughanHumancenteredAgendaIntelligible2021}.

Considering users' knowledge and objectives requires a more nuanced,
context-aware approach to evaluation. Some studies have taken a
bottom-up approach to understanding these needs. Liao et al.
\cite{liaoQuestioningAIInforming2020} interviewed UX and
design practitioners to create an ``XAI question bank'' with
prototypical questions users may wish to have answers to. These include
global questions about \emph{how} a model works, local questions about
\emph{why} a specific prediction was made, counterfactual questions of
\emph{why not} a different prediction, hypothetical questions about
\emph{how to} change the prediction, and more.

A similar approach in education can yield insights into the questions
that teachers, students, and other stakeholders may wish to have
answered by an explanation. Alternatively, it may be that interactive
explanations---perhaps made possible through the abilities of LLMs to
answer questions using natural language---will provide different
stakeholders with the information that is relevant to them, while also
allowing for follow-up questions to better understand explanations
\cite{weldChallengeCraftingIntelligible2019}.

\section{Evaluation criteria}\label{evaluation-criteria}

\begin{figure}
\centering
\includegraphics[width=0.5\linewidth]{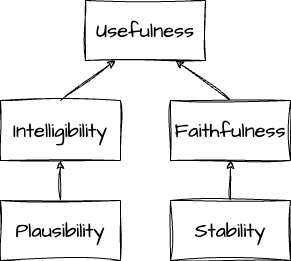}
\caption{Evaluation criteria framework. Edges depict the direction of
dependence (A -\textgreater{} B = A is a prerequisite of
B).\label{framework-diagram}}
\end{figure}

The XAI literature has highlighted several criteria to consider when
evaluating explanations. Due to the lack of a standardized evaluation
framework, these criteria often go by different names, have varying
semantic domains, or are haphazardly used interchangeably. Some of the
definitions used in the literature implicitly suggest the existence of
conceptual dependencies between criteria. However, to the best of our
knowledge, they have not been previously described hierarchically.
Pulling from both the HCI and ML communities, we here propose a
systematic hierarchy of criteria with dependencies between them, as
depicted in Figure \ref{framework-diagram}.

The ultimate goal of an explanation is to be useful to the user. In
education, the user typically represents a stakeholder in the learning
process, such as a teacher, student, parent, or administrator, but it
can also be a researcher who develops and improves the model.

Intuitively, in order for an explanation to be useful, it must meet the
criterion of \emph{intelligibility}, which refers to how well it can be
understood. This concept has also been called ``explicitness''
\cite{alvarezmelisRobustInterpretabilitySelfexplaining2018} and ``comprehensibility''
\cite{carvalhoMachineLearningInterpretability2019}. As discussed in the previous section, the specific context
and target user for which an explanation has been developed is crucial
to an accurate evaluation of intelligibility. In education, an
intelligible explanation is one that can be understood by a student,
teacher, or a different stakeholder, depending on its intended context.
The term ``intelligibility'' arose within the HCI community
\cite{bellottiIntelligibilityAccountabilityHuman2001} and continues to be the predominant term used by HCI
researchers for what the ML community refers to as ``explainability'' or
``interpretability''. We discuss further differences and similarities
between these two communities later in this paper.

Just as intuitively---though slightly more contentiously---useful
explanations must also be faithful. In this context, \emph{faithfulness}
refers to the level of accuracy with which an explanation reflects the
model's internal state
\cite{alvarezmelisRobustInterpretabilitySelfexplaining2018, jacoviFaithfullyInterpretableNLP2020}. Faithful explanations have also been called accurate
explanations \cite{rudinStopExplainingBlack2019}
and high-fidelity explanations
\cite{carvalhoMachineLearningInterpretability2019, nguyenQuantitativeAspectsModel2020}. Faithful explanations can be thought of as providing a
view of the model's internal causality (what leads to its predictions).
In education, a faithful explanation is one that provides accurate
insights into why a model has made a specific content prediction, such
as a study content recommendation or the detection of learner
disengagement.

Unlike with intelligibility, however, there is not universal agreement
on the necessity of explanation faithfulness. This disagreement arises
from the use of post-hoc explainability techniques---such as LIME
\cite{ribeiroWhyShouldTrust2016} and
SHAP \cite{lundbergUnifiedApproachInterpreting2017}---that derive explanations from a simplified approximation
of a more complex model. Post-hoc explanations don't directly access a
model's internal causality, but rather provide a justification of
predictions after-the-fact. Some argue that a lack of faithfulness can
lead to misleading and problematic explanations
\cite{rudinStopExplainingBlack2019, swamyFutureHumancentricEXplainable2023} while others suggest that approximate explanations can be
used to achieve ``sufficient understanding'' for specific users
performing specific tasks
\cite{liaoHumancenteredExplainableAI2022}. We argue that, while perfect faithfulness to a model may not be
necessary in all contexts, a high level of faithfulness is nevertheless
important for an explanation to be useful.

Note that intelligibility and faithfulness are independent criteria. An
explanation can be very intelligible but not particularly faithful, or
highly faithful but quite unintelligible. Yet a useful explanation
requires both conditions to be present past a minimum threshold.

Another criterion described in the literature is \emph{plausibility}
\cite{jacoviFaithfullyInterpretableNLP2020}. A plausible explanation is one that aligns with human intuition.
For example, explaining that a model predicts a student is disengaged
because they did well on a problem is nonsensical. Cases of an
explanation that is faithful but not plausible serve as evidence of a
problem with the model itself---perhaps it is overfitted and is picking
up on noise in the training data. Because plausibility is important to
sensemaking, we argue that it is a prerequisite for intelligibility.

\emph{Stability} refers to the consistency of an explanation for similar
examples
\cite{alvarezmelisRobustInterpretabilitySelfexplaining2018, carvalhoMachineLearningInterpretability2019}. That is, an explanation is stable when it provides similar
results for similar inputs. For example, one would expect a learner
model to provide similar latent knowledge estimates on a particular
knowledge component for students who encountered similar struggles on
the same problems. If an explanation is not stable, it is difficult to
trust it as a reliable source of information. Stability is also a
prerequisite for faithfulness. If an explanation is not adequately
stable, it is unlikely to be faithful to the model's internal state.

\section{Bridging perspectives}\label{bridging-perspectives}

As noted earlier, our evaluation criteria hierarchy for explanations is
informed by two distinct research communities: the ML and HCI
communities. While there is much overlap between them, Liao \& Varshney
\cite{liaoHumancenteredExplainableAI2022} have pointed
out a tension in the goals and methods used by these two communities.
The ML community and the XAI sub-community have primarily focused on
technical solutions to the challenge of interpretability, often relying
on lower-level explainability methods that suit the needs of engineers.
The HCI community, on the other hand, has been more heavily informed by
the social and information sciences, which has led to more user-centered
approaches often based on participatory design methods.

The terms and definitions used by these communities are illustrative of
their differing perspectives. The ML community has settled on terms such
as ``explainability'' and ``interpretability'', and has even fostered a
growing group of \emph{eXplainable} AI (XAI) researchers. The coining of
the term XAI has been attributed to van Lent et al.
\cite{vanlentExplainableArtificialIntelligence2004}, who
used it to describe a system that can present an ``easily understood
chain of reasoning'' from input, ``through the AI's knowledge and
inference'', to the final prediction. The HCI community, on the other
hand, prefers the term ``intelligibility'', which was originally defined
as systems that ``represent to their users what they know, how they know
it, and what they are doing about it''
\cite{bellottiIntelligibilityAccountabilityHuman2001}. Notice the emphasis that explainability places on
the prediction process from input to output, contrasted with the
pragmatic emphasis on users in the HCI definition. Yet despite these
differences, there are more overlaps between these communities than
points of divergence.

It may be that at least part of the tension described by Liao \&
Varshney \cite{liaoHumancenteredExplainableAI2022} is
the result of a \emph{semantic misalignment} between the two groups.
Technical approaches tend to emphasize explanation faithfulness because
they emphasize the role of engineer-researcher as the target user, while
socio-behavioral approaches care more about explanation intelligibility
that have non-researchers as the end-users. In other words, while both
communities are working towards the same goal of making AI
understandable by people, they are doing so from different perspectives
and with different priorities, which leads them to sometimes talk past
each other without realizing it. Vaughan \& Wallach
\cite{vaughanHumancenteredAgendaIntelligible2021} argue
for a bringing together of these communities to create a more holistic
approach to XAI.

It should also be noted that some HCI researchers include aspects of
transparency not often considered to be within the realm of
explainability as crucial to its goals. These go beyond the internal
workings of a model, including explanations of the data used for
training, performance metrics, levels of uncertainty, and the types of
features it relies on
\cite{liaoHumancenteredExplainableAI2022, vaughanHumancenteredAgendaIntelligible2021}. Among education researchers, Kay et al.
\cite{kayScrutableAIED2023} make reference to the
concept of \emph{scrutability} in the sense of being able to scrutinize
a model or system (with a heavy focus on learners as target users). Full
scrutability may require similar aspects of transparency that go beyond
the model itself.

\section{Evaluation methods}\label{evaluation-methods}

An additional layer above that of which evaluation criteria to use is
the question of which methods to use to evaluate explanations. The
evaluation method will dictate the specific criteria that can be
measured. Using Doshi-Velez \& Kim
\cite{doshi-velezRigorousScienceInterpretable2017} as a
guided taxonomy of evaluation methods, we can see how the evaluation
criteria framework we have proposed can be used alongside these
different methods. Within this taxonomy, the choice of evaluation method
depends on the domain-specific needs and the context of intended
interpretability.

Doshi-Velez \& Kim
\cite{doshi-velezRigorousScienceInterpretable2017}
propose three categories of evaluation methods. In decreasing level of
resource complexity, they are:

\begin{itemize}
\tightlist
\item
  \textbf{Application-grounded evaluation}, which involves human users
  performing realistic tasks.
\item
  \textbf{Human-grounded evaluation}, which involves human users
  performing simplified tasks.
\item
  \textbf{Functionally grounded evaluation}, which does not involve
  humans but rather uses quantifiable properties of explanations as a
  proxy for interpretability.
\end{itemize}

An example of application-grounded evaluation in education is the way
learning dashboards are sometimes evaluated by how well they help
instructors understand and provide help to students
\cite{scheersInteractiveExplainableAdvising2021, tissenbaumSupportingClassroomOrchestration2019}, or work on open learner models (OLMs) that provide
students with explanations of a model's estimates of their understanding
\cite{conatiAIEducationNeeds2018}. This
evaluation method can be used to effectively measure explanation
intelligibility (and, by extension, plausibility), but it does not
directly tackle the question of explanation faithfulness.

Functionally grounded evaluation, being the least direct category, makes
it difficult to make any claims about either intelligibility or
faithfulness. It allows for a proxy measurement of intelligibility by
considering properties such as model sparsity or explanation simplicity
\cite{nguyenQuantitativeAspectsModel2020}, but it does not capture the specific needs of any end-user.
While it can be helpful to consider potential target users while
conducting this type of evaluation---ideally realistic stakeholders in
education---the results are generally context-agnostic and therefore may
lack real-world validity. Stability is perhaps the only criterion that
can effectively be evaluated using functionally grounded evaluation.
This method may be most appropriate for preliminary studies in an area
without much prior research.

Some forms of human-grounded evaluation, on the other hand, are more
likely to capture evidence of explanation faithfulness. Doshi-Velez \&
Kim \cite{doshi-velezRigorousScienceInterpretable2017}
identify three examples within this category: binary forced choice,
forward simulation, and counterfactual simulation.

In binary forced choice, participants must select which explanation they
consider best when presented with multiple options. This method was used
in an educational context by Swamy, Du, et al.
\cite{swamyTrustingExplainersTeacher2023} to gauge which
explanations were trusted most by university-level educators. This
somewhat approximates a measurement of plausibility by allowing
participants to identify explanations that match their intuitions, but
it does not truly measure intelligibility. It also does not evaluate
faithfulness.

In forward simulation, participants must correctly predict the model's
output given specific inputs. An experiment along these lines was
proposed by Baker
\cite{bakerChallengesFutureEducational2019} to test
interpretability. This provides a very direct measurement of
faithfulness, since an explanation must be faithful in order for the
task to be performed accurately. It also serves to measure
intelligibility, since participants must understand the explanation to
succeed. However, given a sufficiently simple model, it may be possible
to succeed at a forward simulation task by only using a model's
parameters as explanation without understanding the purpose, features,
or even the domain for which the model was built.

A counterfactual simulation is similar to a forward simulation, but
participants must correctly identify how a specific input needs to be
changed in order to alter the model's given output. This also allows for
an evaluation of both faithfulness and intelligibility, but the same
caveats apply as for forward simulation. The ability to recognize valid
counterfactuals has been identified by Cohausz
\cite{cohauszWhenProbabilitiesAre2022} as a key step
towards using machine learning to design theoretically sound causal
models.

\section{Evaluation case study}\label{evaluation-case-study}

We now turn to an illustration of the concepts discussed here using an
ongoing study of an interpretable neural network for predicting a
particular learner behavior. The model in question is a convolutional
neural network (CNN), designed to be interpretable via targeted
regularization to create binary convolutional filters that more
accurately align with the input data
\cite{jiangPredictiveSequentialPattern2021}. The CNN was trained to predict students' gaming the system
behavior (GTS) on a dataset of interactions with the Cognitive Tutor
Algebra system. The details and results of an early version of this
model were previously reported in Pinto et al.
\cite{pintoInterpretableNeuralNetworks2023}.

\subsection{Setting up the questionnaire
tasks}\label{setting-up-the-questionnaire-tasks}

To evaluate the level of interpretability of this model, we designed a
questionnaire that tasks participants with both a forward simulation and
counterfactual simulation. The questionnaire is designed for
participants from a wide range of backgrounds---both with and without
prior experience in machine learning.

The forward simulation task presents participants with the inputs for a
particular instance---that is, the values for each variable for a given
``clip'' of five consecutive student actions. They must then predict
whether the model would label this clip as \emph{GTS} or \emph{not GTS},
given the patterns in the convolutional filters.

The counterfactual simulation task again presents participants with the
inputs for a specific clip, but this time also providing the model's
predicted label. Participants are asked to identify a single change to
the inputs that would alter the model's prediction. For example, given a
series of inputs and the model's prediction of \emph{not GTS}, what
change to the inputs would result in the model labeling this clip as
\emph{GTS}. Participants select the single correct answer from a series
of multiple-choice options.

\begin{figure}
\centering
\includegraphics[width=1\linewidth]{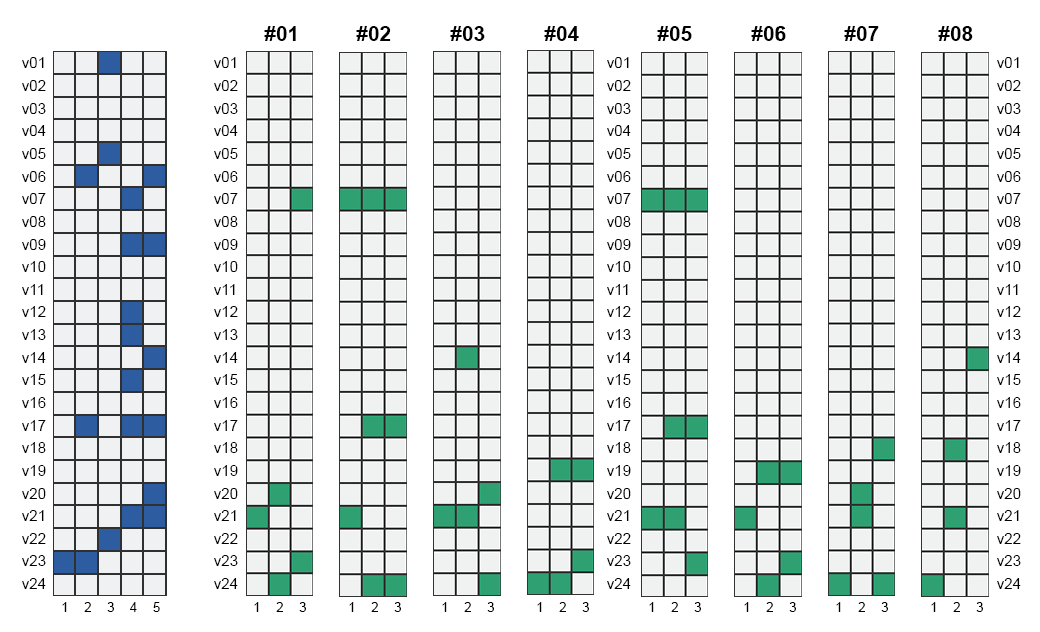}
\caption{Example visualization from the questionnaire that serves as the
core of the model's explanation.\label{questionnaire-sample}}
\end{figure}

Figure \ref{questionnaire-sample} shows an example from the digital
questionnaire platform. The inputs (leftmost blue grid) are presented in
a simplified tabular format: a grid with features stacked vertically
(labeled v01--v24), with each column representing a separate action in
the sequence (labeled 1--5). Blue cells indicate a feature value of 1
(present), while white cells indicate a feature value of 0 (absent). The
binary convolutional filters (green grids) are represented in the same
manner, but each only depicts three actions to match the kernel size of
the model's convolutional layer.

This visualization---along with the background information
provided---serves as the model's explanation, showing the patterns that
the model has learned to associate with GTS behavior. It presents both a
global explanation of the model's logic (patterns that are indicative of
GTS) and explanations of specific outputs (the model's prediction for a
particular clip of student actions). The questionnaire is used as a tool
to evaluate the explanations themselves.

For the forward simulation task, the instructions ask the following
questions: ``would the model identify the following clip of student
actions as GTS or not GTS? If GTS, what is the number of the matching
model pattern?'' In the counterfactual simulation task, we ask ``which
of the following changes to the input would alter the model's
prediction?'' Possible answers for the counterfactual simulation include
the addition or removal of specific actions, such as ``add v07 to action
2'' or ``remove v19 at action 4''. For both tasks, we also ask
participants to rate their confidence on each question.

\subsection{Evaluating the evaluation}\label{evaluating-the-evaluation}

The evaluation methods used in this questionnaire clearly fall within
the category of human-grounded evaluation in the method taxonomy
proposed by Doshi-Velez \& Kim
\cite{doshi-velezRigorousScienceInterpretable2017}---they
involve human users performing simplified tasks. As such, they provide
measurements of faithfulness and intelligibility. By calculating the
average accuracy rate (proportion correct out of total questions) across
the entire sample of participants, we can quantify how well the
explanations were understood (intelligibility). Because the tasks align
so closely with the model's actions, the accuracy rate also serves to
measure how well the explanations reflect the model's internal state
(faithfulness).

However, the caveat provided earlier in regards to forward and
counterfactual simulation tasks applies here---participants are not
required to understand the specific purpose of the model or the value of
its predictions in order to succeed. In fact, while we present an
explanation of GTS and the aims of the model as background information,
we've entirely excluded meaningful feature labels from the explanations.
This approach makes it impossible to evaluate explanation plausibility,
weakening its claims of evaluating intelligibility beyond a
surface-level understanding.

Furthermore, this questionnaire does not claim to target any specific
end-users. It has been designed for participants from a wide range of
backgrounds, and for no specific purpose other than its completion. We
previously highlighted the importance of intended context when
evaluating explanations, which is difficult to account for using the
simplified tasks of human-grounded methods. An application-grounded
evaluation would allow for a better understanding of the specific needs
of end-users, but it would also make it difficult to measure
faithfulness and would require a more complex and time-consuming study
design
\cite{doshi-velezRigorousScienceInterpretable2017}. When it comes to designing an evaluation, tradeoffs may
be necessary.

\section{Discussion}\label{discussion}

Much like the complexity of evaluating the different aspects of a
model's performance, the evaluation of explanations is itself a complex
task and cannot be captured in its entirety by any one metric or method.
We have aimed to provide an initial framework to guide this daunting but
important task. However, much work remains to be done.

We have brought together evaluation criteria described by different
communities into a cohesive whole, but they largely remain abstract
ideas. In order to be useful in practice, these criteria must be
operationalized more concretely in the educational contexts in which we
wish to use them. Furthermore, this high-level overview is likely
missing key criteria that measure aspects of explanations that are
currently not being captured.

For example, when describing the aspects of intelligibility that can be
captured by human-grounded evaluation methods, as well as those that may
go overlooked by such an approach, we found that we didn't have the
exact language to elucidate our point. It may be that there is an
element of intelligibility that requires an additional criterion to
fully capture---something along the lines of an explanation's fidelity
to its intended context.

Similarly, the framework's hierarchical structure itself may benefit
from further scrutiny. Edge cases theoretically could exist that don't
perfectly fit, such as the possibility of a highly overfitted model
leading to explanations that are faithful but not very stable.

Nevertheless, future research may build on the framework and ideas
presented here to create a more comprehensive evaluation framework for
explanations. A unified framework should be adaptable to the specific
needs of different contexts, should be informed by the perspectives of
both the technical ML and human-centered HCI communities, and should be
relevant to the needs of stakeholders in education.

\section{Conclusion}\label{conclusion}

In this position paper, we have proposed the need for a unified
framework for evaluating explanations in the context of XAI. We have
reviewed important concepts for better understanding the nature of
explanations, including their role as mediators between models and
users, the central role played by an explanation's intended context, and
the varied perspectives brought by different research communities. We
have further argued that useful explanations require both faithfulness
and intelligibility, and have proposed a hierarchy of criteria that
brings together concepts previously described in the literature.
Finally, we have provided a case study for these criteria using the
ongoing evaluation of a neural-network-based learner behavior detector.

\bibliographystyle{abbrv}
\bibliography{sigproc.bib}

\balancecolumns
\end{document}